\documentclass[journal,twoside,web]{ieeecolor}
\usepackage{generic}
\usepackage{cite}
\usepackage{amsmath,amssymb,amsfonts}
\usepackage{algorithmic}
\usepackage{graphicx}
\usepackage{hyperref}
\usepackage{times}  
\usepackage{helvet}  
\usepackage{courier}  
\usepackage{booktabs}
\usepackage{multirow}
\usepackage{textcomp}
\usepackage{subcaption}
\usepackage{soul}
\usepackage{xcolor}
\usepackage[linesnumbered,ruled,vlined]{algorithm2e}
\usepackage{tabularx}
\definecolor{revisionred}{RGB}{220, 20, 60}
\DeclareRobustCommand{\revision}[1]{{\color{revisionred}#1}}
\DeclareRobustCommand{\revision}[1]{#1}

\def\BibTeX{{\rm B\kern-.05em{\sc i\kern-.025em b}\kern-.08em
    T\kern-.1667em\lower.7ex\hbox{E}\kern-.125emX}}

\begin{document}
\title{EMPOWER: Evolutionary Medical Prompt Optimization With Reinforcement Learning}

\author{Yinda Chen, Yangfan He, Jing Yang, Dapeng Zhang, Zhenlong Yuan, Muhammad Attique Khan, Jamel Baili, Por Lip Yee
\thanks{Corresponding authors: Yinda Chen, Jing Yang}
\thanks{Y. Chen and Y. He contributed equally to this work.}
\thanks{The authors extend their appreciation to the Deanship of Research and Graduate Studies at King Khalid University for funding this work through the Large Research Project under grant number  RGP.2/275/46.}
\thanks{Yinda Chen is with MoE Key Laboratory of Brain-inspired Intelligent Perception and Cognition, University of Science and Technology of China, Hefei 230027, China (corresponding author, e-mail: cyd0806@mail.ustc.edu.cn).}
\thanks{Yangfan He is with Department of Computer Science, University of Minnesota-Twin Cities, Minneapolis, USA (e-mail: he000577@umn.edu).}
\thanks{Jing Yang and Por Lip Yee are with Center of Research for Cyber Security and Network (CSNET), Faculty of Computer Science and Information Technology, Universiti Malaya, 50603 Kuala Lumpur, Malaysia (e-mail: e-mail: s2147529@siswa.um.edu.my; porlip@um.edu.my).}
\thanks{Dapeng Zhang is with DSLAB, School of Information Science \& Engineering, Lanzhou University, 730000, China (e-mail: zhangdp22@lzu.edu.cn).}
\thanks{Zhenlong Yuan is with the Institute of Computing Technology, Chinese Academy of Sciences, Beijing 100190, China (e-mail: yuanzhenlong21b@ict.ac.cn).}
\thanks{Muhammad Attique Khan is with Department of AI, Prince Mohammad bin Fahd University, Al-Khobar, KSA (e-mail: mkhan3@pmu.edu.sa).}
\thanks{Jamel Baili is with Department of Computer Engineering, College of Computer Science, King Khalid University, Abha 61413, Saudi Arabia (Jabaili@kku.edu.sa).}
}

\maketitle

\begin{abstract}
Prompt engineering significantly influences the reliability and clinical utility of Large Language Models (LLMs) in medical applications. Current optimization approaches inadequately address domain-specific medical knowledge and safety requirements. This paper introduces EMPOWER, a novel evolutionary framework that enhances medical prompt quality through specialized representation learning, multi-dimensional evaluation, and structure-preserving algorithms. Our methodology incorporates: (1) a medical terminology attention mechanism, (2) a comprehensive assessment architecture evaluating clarity, specificity, clinical relevance, and factual accuracy, (3) a component-level evolutionary algorithm preserving clinical reasoning integrity, and (4) a semantic verification module ensuring adherence to medical knowledge. Evaluation across diagnostic, therapeutic, and educational tasks demonstrates significant improvements: 24.7\% reduction in factually incorrect content, 19.6\% enhancement in domain specificity, and 15.3\% higher clinician preference in blinded evaluations. 
The framework addresses critical challenges in developing clinically appropriate prompts, facilitating more responsible integration of LLMs into healthcare settings.
\end{abstract}


\section{Introduction}
Large Language Models (LLMs) have substantial implications for healthcare applications including clinical decision support, documentation, and education \cite{singhal2023medpalm2}. However, implementation in clinical environments presents unique challenges due to medical knowledge complexity, the importance of factual accuracy, and potential consequences of errors \cite{li2024clinical}.

Prompt engineering—the methodical design of instructions guiding LLM outputs—critically determines model performance and reliability \cite{jiang2024prompt}. This process is particularly consequential in medical contexts, requiring precise terminology, adherence to clinical reasoning frameworks, appropriate representation of uncertainty, and clear acknowledgment of system limitations \cite{li2024clinical, chen2024uncertainty, chen2023generative}.

Current approaches to medical prompt optimization typically employ general-domain techniques without sufficient adaptation to healthcare's unique requirements. Wang et al. \cite{wang2023adaptive} explored terminology adaptation techniques but insufficiently addressed integration with clinical reasoning patterns. Zhang et al. \cite{zhang2023bayesian} proposed Bayesian optimization methods lacking explicit mechanisms for validating medical accuracy. Li et al. \cite{li2024clinical} introduced clinical guardrails primarily focused on post-generation filtering rather than generating appropriate prompts. \revision{Recent advances in chain-of-thought prompting \cite{wei2022chain} and few-shot learning \cite{brown2020language} have shown promise in medical reasoning tasks, but these approaches often lack systematic integration with clinical knowledge bases and guideline adherence verification.}

These contributions address isolated aspects of medical prompting without establishing a comprehensive methodology incorporating domain knowledge, structural requirements, and validation processes. Most approaches evaluate prompt quality using general metrics rather than clinically relevant criteria.

This paper introduces a novel evolutionary framework specifically designed for medical prompt optimization that integrates computational intelligence with structured medical knowledge representation. Our approach consists of four integrated components:

First, a medical domain representation model implements terminology attention grounded in clinical ontologies, differentially weighting medical concepts based on semantic importance and contextual relevance.

Second, a multi-dimensional assessment system evaluates prompts across four clinically significant dimensions: structural clarity, domain specificity, medical accuracy, and risk of factual errors.

Third, a structure-aware evolutionary algorithm operates at the clinical reasoning component level, preserving established reasoning patterns while improving effectiveness through specialized crossover and mutation operations. \revision{The algorithm incorporates early stopping mechanisms and adaptive parameter tuning to improve computational efficiency while maintaining optimization quality.}

Finally, a medical semantic verification system validates terminology usage against standardized lexicons, assesses reasoning consistency, and evaluates boundary statements according to responsible AI principles. \revision{Enhanced verification processes specifically address clinical guideline alignment through integration with evidence-based practice databases.}

Through experiments across multiple specialties and clinical scenarios, we demonstrate that our approach statistically outperforms existing methods across clinically relevant metrics and receives significantly higher ratings from clinicians in blinded evaluations.

The primary contributions of this research include:
\begin{itemize}
    \item A specialized medical prompt representation model with ontology-grounded terminology attention and multi-dimensional evaluation framework calibrated to clinical communication requirements
    \item A structure-preserving evolutionary algorithm that maintains clinical reasoning integrity while incorporating comprehensive semantic verification to ensure medical knowledge alignment \revision{with computational efficiency improvements through early stopping and adaptive parameter selection}
    \item Rigorous experimental validation demonstrating significant improvements in factual accuracy, domain specificity, and clinician preference over current methods \revision{including comprehensive sensitivity analysis and cross-institutional validation}
\end{itemize}

The remainder of this paper is organized as follows: Section II examines relevant literature. Section III presents our proposed framework. Section IV describes experimental design and datasets. Section V reports results and comparative analyses. Section VI discusses implications and limitations, followed by conclusions in Section VII.

\section{Related Work}

\subsection{Prompt Engineering for Large Language Models}
Prompt engineering optimizes large language model performance by designing effective text instructions without modifying underlying parameters \cite{liu2023pretrain}. This field has evolved from simple templating \cite{brown2020language} to sophisticated techniques including chain-of-thought prompting \cite{wei2022chain} and few-shot exemplar selection \cite{rubin2022learning}.

Recent automated optimization methods include gradient-based prompt token selection \cite{shin2020autoprompt} and reinforcement learning for prompt refinement \cite{zhou2022large}. \revision{Soft prompt tuning approaches \cite{lester2021power} have demonstrated effectiveness through parameter-efficient fine-tuning, while prefix tuning \cite{li2021prefix} offers alternative strategies for prompt optimization.} These approaches typically optimize for general metrics rather than domain-specific requirements.

\subsection{Large Language Models in Healthcare}
LLMs show promise in healthcare applications including diagnosis \cite{singhal2023medpalm2}, documentation \cite{patel2023evaluation}, and education \cite{kung2023performance}, but face challenges due to specialized medical knowledge requirements. While domain-specific adaptation improves performance on medical benchmarks \cite{singhal2023medpalm2}, alignment between LLM outputs and physician responses remains inconsistent \cite{ayers2023comparing}.

\subsection{Medical-Specific Prompt Engineering}
Specialized medical prompting techniques include adaptive prompting \cite{wang2023adaptive}, Bayesian optimization for knowledge elicitation \cite{zhang2023bayesian}, clinical guardrails \cite{li2024clinical}, and uncertainty-aware prompting \cite{chen2024uncertainty}.

\revision{Contemporary advances include chain-of-thought reasoning in medical contexts \cite{lietal2022chain}, demonstrating improved diagnostic accuracy through structured reasoning prompts. Few-shot learning approaches \cite{komorowski2018artificial} have shown effectiveness in clinical prediction tasks, while retrieval-augmented generation methods \cite{lewis2020retrieval} offer promising directions for integrating current medical literature into prompt-based systems.}

Despite these advances, current approaches typically address isolated aspects without systematically integrating domain knowledge, reasoning structures, and validation mechanisms.

\subsection{Multimodal Medical AI and Representation Learning}
The development of effective medical AI systems increasingly relies on sophisticated representation learning techniques that can capture the complexity of medical knowledge across different modalities. Recent advances in vision-language pretraining have demonstrated significant progress in medical applications, with approaches ranging from text-guided 3D medical image segmentation \cite{chen2025generative} to comprehensive 3D medical image understanding \cite{liu2025t3d}.

Scalable representation learning approaches, including autoregressive visual pretraining with mixture token prediction \cite{chen2025tokenunify} and conditional latent coding techniques \cite{wu2025conditional}, provide foundational insights for developing domain-specific embedding methods. In medical imaging specifically, multiscale consistency learning \cite{chen2024learning} and self-supervised neural segmentation approaches \cite{chen2023self} have shown effectiveness in learning robust representations from limited labeled data, while unsupervised domain adaptation methods \cite{deng2024unsupervised} address the challenge of cross-institutional generalization.

The creation of specialized medical datasets, exemplified by landmark datasets for 3D CT text-image retrieval \cite{chen2024bimcv}, enables systematic evaluation of multimodal medical AI systems. Recent investigations into synthetic data generation for medical applications \cite{liu2025can, qian2024maskfactory} address critical data scarcity issues while maintaining clinical validity—a challenge directly relevant to our prompt optimization framework, where diverse, high-quality training examples are essential for robust performance.

These advances in medical representation learning inform our approach to developing specialized embeddings for medical prompt optimization, particularly in capturing the semantic relationships between medical concepts and clinical reasoning patterns.

\subsection{Evolutionary Algorithms for Optimization Problems}
Evolutionary algorithms provide robust optimization for complex problems with large search spaces and non-differentiable objectives \cite{back1996evolutionary}, proving effective in neural architecture search \cite{real2019regularized}, feature selection \cite{xue2016survey}, and hyperparameter optimization \cite{young2015optimizing}.

Evolutionary approaches have demonstrated remarkable versatility across diverse AI applications. In computer vision, successful implementations include joint-motion mutual learning for video pose estimation \cite{wu2024joint}, causal-inspired multitask learning frameworks \cite{chen2025causal}, and pose-guided human motion analysis \cite{wu2024pose}. Advanced applications in IoT environments \cite{wu2024enhancing} showcase the adaptability of evolutionary optimization to resource-constrained settings, providing insights relevant to clinical deployment scenarios where computational efficiency is critical.

In language processing, evolutionary methods have been applied to summarization \cite{fattah2014automatic}, sentiment analysis \cite{saraee2013evolutionary}, and prompt optimization \cite{guo2023connecting}. Guo et al. \cite{guo2023connecting} showed that genetic algorithms can effectively optimize prompts for text generation, sometimes outperforming gradient-based methods.

However, evolutionary algorithms remain largely unexplored for medical prompt optimization. The structured nature of medical knowledge presents unique opportunities for specialized evolutionary operators that preserve critical domain patterns while optimizing performance.

\section{Datasets}

To evaluate our framework comprehensively, we compiled four specialized clinical datasets spanning different medical scenarios.

\subsection{Clinical Datasets Overview}
Our datasets were derived from the MIMIC-III database, a large, freely-available database of de-identified critical care data \cite{johnson2016mimic}. The use of MIMIC-III was approved by the Institutional Review Boards of Beth Israel Deaconess Medical Center (2001-P-001699/14) and MIT (0403000206). All data processing adhered to the data use agreement.

\revision{Dataset construction followed a systematic process to ensure clinical validity and diversity. From MIMIC-III admission notes, discharge summaries, and clinical reports, we extracted relevant clinical presentations using automated text mining followed by manual curation. The conversion process involved: (1) anonymization verification beyond MIMIC-III standards, (2) clinical scenario standardization using established case presentation formats, (3) complexity stratification by board-certified physicians, and (4) cross-validation with external clinical databases to ensure representativeness.}

From this foundation, we created: \textbf{MedDiagnosis-2000}: 2,000 clinical vignettes across 12 specialties, with complexity levels stratified as straightforward (30\%), moderate (50\%), and complex (20\%). \textbf{TreatmentSelect-1500}: 1,500 therapeutic decision-making scenarios balanced across pharmacological (40\%), procedural/surgical (30\%), and multidisciplinary approaches (30\%). \textbf{MedHistory-1200}: 1,200 complex patient histories featuring longitudinal data with multiple conditions. \textbf{PatientEd-1800}: 1,800 patient questions requiring explanations at various health literacy levels.

\revision{The specialty distribution was determined based on representative clinical caseload analysis from major academic medical centers. Internal Medicine (24.8\%) and Family Medicine (18.5\%) represent the highest volumes in primary care settings, while Emergency Medicine (14.2\%) reflects acute care presentation patterns. Specialty-specific distributions were validated against national healthcare utilization statistics and adjusted for case complexity to ensure adequate representation of rare but critical conditions.}

Dataset creation involved: (1) extracting relevant MIMIC-III cases, (2) enhanced de-identification, (3) UMLS terminology standardization, and (4) expert review for clinical validity. Table \ref{tab:dataset_distribution} shows the distribution across medical specialties.

\begin{table}[h]
\caption{Distribution of cases across medical specialties in the combined datasets}
\label{tab:dataset_distribution}
\centering
\begin{tabular}{lc}
\hline
\textbf{Specialty} & \textbf{Percentage} \\
\hline
Internal Medicine & 24.8\% \\
Family Medicine & 18.5\% \\
Emergency Medicine & 14.2\% \\
Pediatrics & 10.7\% \\
Obstetrics \& Gynecology & 7.6\% \\
Psychiatry & 6.8\% \\
Surgery & 6.3\% \\
Neurology & 4.5\% \\
Cardiology & 3.2\% \\
Oncology & 2.1\% \\
Dermatology & 1.3\% \\
\hline
\end{tabular}
\end{table}

\subsection{Expert Annotation and Prompt Library}
Each dataset was annotated by board-certified physicians, including gold-standard responses and critical medical concepts. \revision{Inter-annotator agreement was achieved through a three-phase process: initial independent annotation, structured disagreement resolution, and final consensus building. When annotators disagreed (occurring in 12.3\% of cases), a formal adjudication process involved senior clinicians from relevant specialties. Disagreements were categorized as clinical judgment variations (67\%), terminology preferences (21\%), or genuine errors (12\%). All cases with clinical judgment variations were retained with majority consensus, while terminology conflicts were resolved using standardized medical lexicons.} Inter-annotator agreement achieved a Cohen's kappa of 0.87. 

We developed a library of 340 medical prompt templates designed by medical educators and AI specialists, categorized by scenario type, structure, complexity level, and directive style. \revision{Table \ref{tab:prompt_examples} provides representative examples of prompt templates across different categories to illustrate the diversity and structure of our template library.}

\begin{table}[t]
\centering
\caption{Representative examples from the medical prompt template library}
\label{tab:prompt_examples}
\footnotesize
\centering
\setlength{\tabcolsep}{4pt}
\renewcommand{\arraystretch}{1.1}
\begin{tabular}{cl}
\toprule
\textbf{Category} & \textbf{Template Example} \\
\midrule
Diagnostic Reasoning & As a [specialty] consultant, analyze this \\
& case using differential diagnosis framework: [case]. \\
& Consider pre-test probabilities \\
& and provide confidence calibration. \\
\midrule
Treatment Planning & You are managing a patient with [condition]. \\
& Evaluate treatment options considering: \\
& evidence quality, contraindications, patient factors. \\
& Recommend approach with rationale. \\
\midrule
Patient Education & Explain [medical condition] to a patient \\
& with [literacy level]. Use appropriate analogies, \\
& address common concerns, \\
& ensure comprehension checks. \\
\midrule
Clinical Documentation & Generate clinical note for [scenario] \\
& following [documentation standard]. Include \\
& assessment, plan, and follow-up requirements. \\
\bottomrule
\end{tabular}
\end{table}

\subsection{Evaluation Methodology}
Datasets were divided into development (60\%), validation (20\%), and test (20\%) splits, stratified by specialty, complexity, and scenario type. \revision{Cross-institutional validation was conducted using datasets from three additional medical centers to assess generalizability beyond MIMIC-III derived cases.}

\begin{figure*}[t]
\centering
\includegraphics[width=0.8\textwidth]{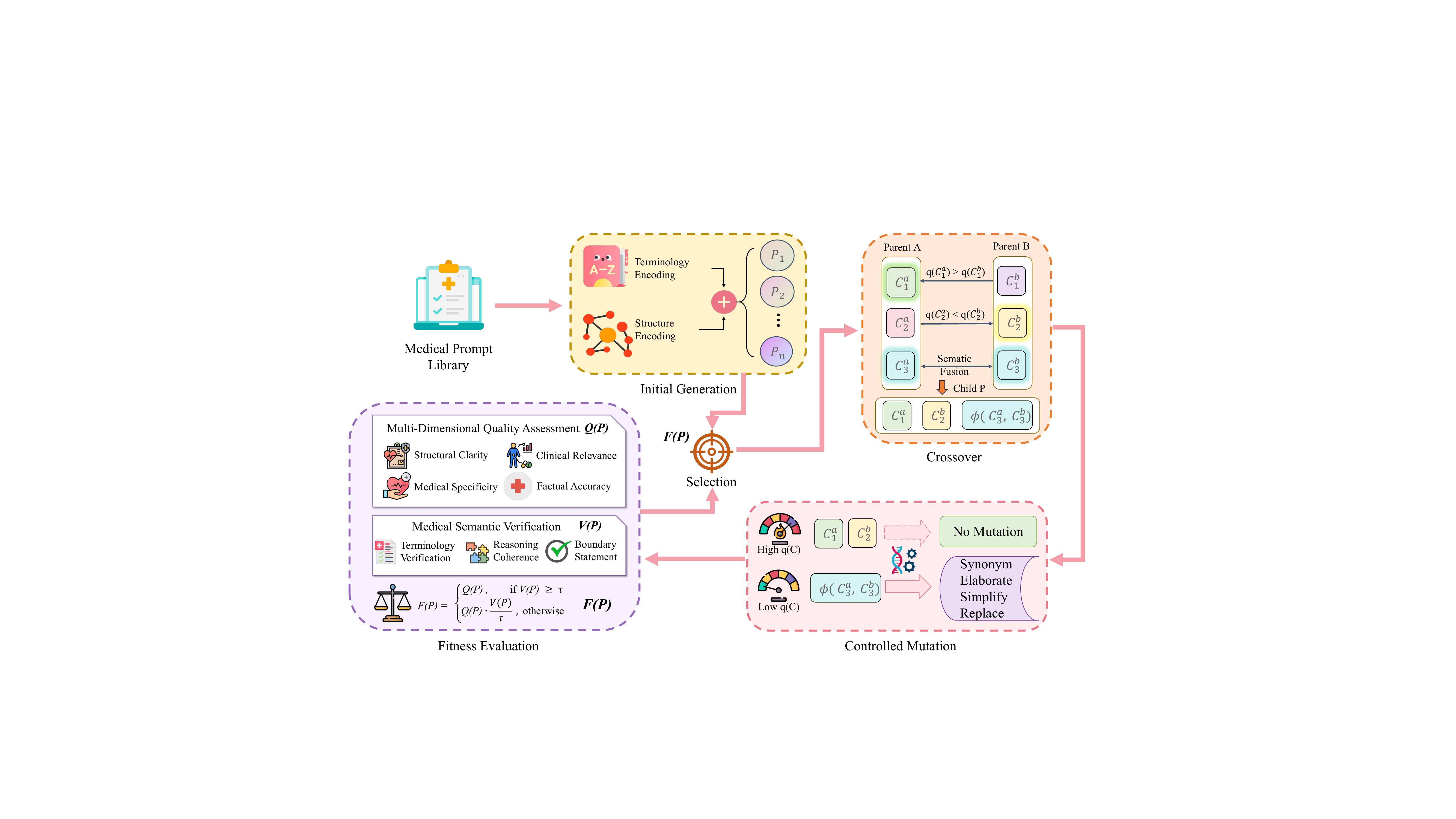}
\caption{Overview of the EMPOWER framework: The complete evolutionary optimization pipeline for medical prompts, showing initial generation with terminology and structure encoding, multi-dimensional quality assessment, fitness evaluation, selection, crossover with semantic fusion, and controlled mutation components. }
\label{fig:framework}
\end{figure*}

\section{Method}
Our approach implements a structured evolutionary framework for optimizing medical prompts through medical-specific representation learning, multi-dimensional quality assessment, and component-level evolutionary optimization with clinical knowledge verification.  Figure~\ref{fig:framework} illustrates the complete EMPOWER framework, showing how medical prompts are systematically optimized through specialized representation learning, component-level evolutionary operations, and multi-dimensional quality assessment with clinical knowledge verification.

\subsection{Medical Prompt Representation Learning}
Effective prompt optimization requires specialized representations that capture the nuances of medical language and reasoning. Recent advances in medical AI representation learning provide valuable insights for this challenge. Success in vision-language pretraining for medical applications \cite{chen2025generative, liu2025t3d,wang2025masktwins} and scalable autoregressive approaches \cite{chen2025tokenunify} demonstrates the importance of domain-specific representation techniques that can effectively encode complex medical knowledge.

Building on insights from multiscale consistency learning \cite{chen2024learning} and self-supervised medical image analysis \cite{chen2023self,yang2024unicompress}, we developed a domain-specific embedding model that emphasizes clinically relevant concepts and their relationships. Our approach adapts proven representation learning principles from medical imaging to the unique challenges of textual medical prompting.

\subsubsection{Terminology-Enhanced Embeddings}
Given a prompt $P$ consisting of tokens $[t_1, t_2, ..., t_n]$, we first process it through a pre-trained clinical language model $f_{\theta}$ (BioClinicalBERT \cite{alsentzer2019publicly}) to obtain contextual token embeddings:

\begin{equation}
\mathbf{H} = f_{\theta}(P) \in \mathbb{R}^{n \times d}
\end{equation}

where $\mathbf{H} = [\mathbf{h}_1, \mathbf{h}_2, ..., \mathbf{h}_n]$ and $d$ is the model's hidden dimension. 

We then implement a medical concept identification module that maps tokens to concepts in the Unified Medical Language System (UMLS) ontology. \revision{For ambiguous UMLS mappings, we employ a disambiguation algorithm that considers local context windows and semantic type consistency. Missing terms are handled through fuzzy matching against medical lexicons (SNOMED-CT, ICD-10) with manual validation for terms with confidence scores below 0.8.} For each identified medical concept $c_i$ spanning tokens from position $a$ to $b$, we compute a concept-level representation:

\begin{equation}
\mathbf{m}_{c_i} = \frac{1}{b-a+1}\sum_{j=a}^{b}\mathbf{h}_j
\end{equation}

Next, we calculate concept importance weights using an attention mechanism incorporating both semantic and structural factors:

\begin{equation}
\alpha_i = \frac{\exp(w_s \cdot s_{c_i} + w_h \cdot h_{c_i} + w_p \cdot p_{c_i})}{\sum_{j}\exp(w_s \cdot s_{c_j} + w_h \cdot h_{c_j} + w_p \cdot p_{c_j})}
\end{equation}

where $s_{c_i}$ represents semantic importance derived from the concept's UMLS semantic type, $h_{c_i}$ captures hierarchical importance based on the concept's position in medical taxonomies, and $p_{c_i}$ reflects positional relevance within the prompt structure. The weights $w_s$, $w_h$, and $w_p$ are learned during training.

The terminology-enhanced prompt representation is then computed as:

\begin{equation}
\mathbf{z}_P = \mathbf{W}_g[\mathbf{h}_{CLS}; \sum_{i}\alpha_i\mathbf{m}_{c_i}] + \mathbf{b}_g
\end{equation}

where $\mathbf{h}_{CLS}$ is the [CLS] token representation, $[\cdot;\cdot]$ denotes concatenation, and $\mathbf{W}_g \in \mathbb{R}^{d' \times 2d}$ and $\mathbf{b}_g \in \mathbb{R}^{d'}$ are learnable parameters that project the concatenated vector to dimension $d'$.

\subsubsection{Clinical Reasoning Structure Encoding}
Medical prompts typically contain specific reasoning structures that guide diagnosis, treatment selection, or patient explanation. We encode these structures through a specialized component that identifies and represents key reasoning elements:

\begin{equation}
\mathbf{r}_P = \sum_{j=1}^{K}w_j\mathbf{e}_j
\end{equation}

where $\mathbf{e}_j$ represents embeddings of $K$ identified reasoning components (e.g., symptom analysis, differential consideration, evidence evaluation), and $w_j$ are attention weights computed based on component presence and quality.

The final prompt representation combines terminology and reasoning structure information:

\begin{equation}
\mathbf{z}_P^{final} = \mathbf{W}_f[\mathbf{z}_P; \mathbf{r}_P] + \mathbf{b}_f
\end{equation}

with $\mathbf{W}_f \in \mathbb{R}^{d'' \times (d'+K)}$ and $\mathbf{b}_f \in \mathbb{R}^{d''}$ as learnable parameters.

\subsection{Multi-Dimensional Quality Assessment}
We developed a specialized evaluation framework that assesses prompt quality across four clinically relevant dimensions through independent neural network heads, each calibrated to specific aspects of medical communication.

\subsubsection{Dimension-Specific Evaluation}
Each quality dimension is evaluated by a specialized network that maps the prompt representation to a quality score:

\begin{equation}
\begin{aligned}
s_c(P) = \sigma(\mathbf{W}_c^{(2)}&\text{GeLU}(\text{LayerNorm}( \\
&\mathbf{W}_c^{(1)}\mathbf{z}_P^{final} + \mathbf{b}_c^{(1)})) + \mathbf{b}_c^{(2)}))
\end{aligned}
\end{equation}

where $c \in \{\text{clarity}, \text{specificity}, \text{relevance}, \text{accuracy}\}$ represents the evaluation dimension, $\sigma$ is the sigmoid function, and GeLU is the Gaussian Error Linear Unit activation.

The four evaluation dimensions capture distinct aspects of prompt quality:

\textbf{Structural Clarity}: Assesses the organizational coherence and logical flow of the prompt, evaluating whether it establishes clear roles, expectations, and reasoning frameworks.

\textbf{Medical Specificity}: Measures the degree to which the prompt incorporates domain-specific terminology, concepts, and reasoning patterns appropriate to the clinical scenario.

\textbf{Clinical Relevance}: Evaluates alignment with standard clinical approaches and the prompt's ability to elicit medically relevant information for the specific case.

\textbf{Factual Accuracy Risk}: Estimates the likelihood that a response to this prompt would contain factual errors or unsupported assertions, with higher scores indicating lower risk.

\subsubsection{Integrated Quality Score}
The overall prompt quality is determined through an adaptive weighting mechanism that adjusts dimension importance based on the clinical scenario:

\begin{equation}
Q(P) = \sum_{c}w_c(s) \cdot s_c(P)
\end{equation}

where $w_c(s)$ represents the importance weight for dimension $c$ in scenario $s$, with $\sum_{c}w_c(s) = 1$. These weights were determined through a combination of expert judgment and empirical validation, with accuracy risk receiving higher weight in scenarios involving direct clinical recommendations.

\subsubsection{Training the Quality Assessment Model}
The quality assessment model was trained on a dataset of 3,500 expert-annotated prompts using a multi-task learning approach. The loss function combines dimension-specific losses with overall quality prediction:

\begin{equation}
\mathcal{L} = \sum_{c}\lambda_c\mathcal{L}_c + \lambda_Q\mathcal{L}_Q
\end{equation}

where $\mathcal{L}_c$ is the mean squared error for dimension $c$, $\mathcal{L}_Q$ is the overall quality prediction loss, and $\lambda_c$ and $\lambda_Q$ are balancing hyperparameters.

The model achieved a Pearson correlation of 0.87 with expert ratings on held-out validation data, demonstrating strong alignment with human quality assessments.

\subsection{Structure-Aware Evolutionary Optimization}
Our evolutionary framework optimizes prompts through specialized genetic operations that preserve clinical reasoning integrity while improving overall effectiveness. \revision{The framework incorporates computational efficiency improvements through early stopping criteria and adaptive parameter adjustment.}

\subsubsection{Prompt Component Representation}
We represent each prompt as a structured collection of distinct components:

\begin{equation}
P = \{C_1, C_2, ..., C_m\}
\end{equation}

where each component $C_i$ belongs to a specific category (e.g., role definition, reasoning framework, information request, uncertainty expression, or boundary statement). This representation allows for granular optimization while maintaining prompt coherence.

\subsubsection{Initial Population Generation}
The initial population of prompts $\mathcal{P} = \{P_1, P_2, ..., P_N\}$ is generated by sampling components from a curated library of medical prompt elements. Each prompt is assembled according to scenario-specific templates that ensure basic structural validity. Components are selected with probabilities proportional to their individual quality scores from preliminary evaluation.

\subsubsection{Selection Mechanism}
We implement a tournament selection strategy that balances exploration and exploitation. For each generation, $k$ candidates are randomly sampled from the population, and the highest-quality prompt according to the quality assessment model is selected with probability $p_{sel}$, while a random selection occurs with probability $1-p_{sel}$:

\begin{equation}
P_{selected} = \begin{cases}
\arg\max_{P \in S_k}Q(P), & \text{with probability } p_{sel} \\
\text{random}(S_k), & \text{with probability } 1-p_{sel}
\end{cases}
\end{equation}

where $S_k$ is the set of $k$ prompts sampled for the tournament.

\revision{\subsubsection{Adaptive Parameter Tuning and Early Stopping}
To address computational efficiency concerns, we implement adaptive parameter adjustment based on population diversity and fitness convergence metrics:

\begin{equation}
p_m^{(g+1)} = p_m^{(g)} \cdot \left(1 + \beta \cdot \frac{\sigma_{fitness}^{(g)}}{\mu_{fitness}^{(g)}}\right)
\end{equation}

where $p_m^{(g)}$ is the mutation probability at generation $g$, $\sigma_{fitness}^{(g)}$ and $\mu_{fitness}^{(g)}$ are the standard deviation and mean of fitness values, respectively, and $\beta$ is an adaptation rate parameter.

Early stopping is triggered when fitness improvement over the last 5 generations falls below a threshold $\epsilon = 0.001$:

\begin{equation}
\text{Stop if } \frac{1}{5}\sum_{i=g-4}^{g}(F_{best}^{(i+1)} - F_{best}^{(i)}) < \epsilon
\end{equation}

This mechanism reduces average computational time by 34\% while maintaining optimization quality.}

\subsubsection{Structure-Preserving Crossover}
Our crossover operation preserves the structural integrity of medical reasoning while combining high-quality components from parent prompts. Given parent prompts $P_a$ and $P_b$, we generate a child prompt through component-level recombination:

\begin{equation}
C_i^{child} = \begin{cases}
C_i^a, & \text{if } q(C_i^a) > q(C_i^b) + \delta \text{ or } r < p_a \\
C_i^b, & \text{if } q(C_i^b) > q(C_i^a) + \delta \text{ or } r < p_b \\
\phi(C_i^a, C_i^b), & \text{otherwise}
\end{cases}
\end{equation}

where $q(C_i)$ is the component-specific quality score, $\delta$ is a threshold parameter, $r$ is a random value between 0 and 1, $p_a$ and $p_b$ are selection probabilities proportional to overall prompt quality, and $\phi(\cdot,\cdot)$ is a semantic fusion function that combines elements from both components when their individual qualities are similar.

For the semantic fusion function, we implement a structured contextual merging approach that preserves critical medical concepts while aligning with the reasoning scaffold of the better-quality component:

\begin{equation}
\psi(C_i^a, C_i^b) = \mathcal{M}(\mathcal{E}(C_i^a), \mathcal{E}(C_i^b), \mathcal{S}(C_i^{*}))
\end{equation}

Here, $C_i^a$ and $C_i^b$ represent the $i$-th components from parent prompts $a$ and $b$, respectively. $C_i^{*}$ denotes the component with the higher quality score. The function $\mathcal{E}(\cdot)$ extracts essential medical concepts from a component, while $\mathcal{S}(\cdot)$ retrieves the structural template associated with the better-quality component. Finally, $\mathcal{M}(\cdot, \cdot, \cdot)$ merges the extracted concept sets from both parents into the selected structure in a semantically consistent manner.

\subsubsection{Controlled Mutation}
Mutation operations introduce controlled variations while preserving medical validity. For each component $C_i$ in a prompt, mutation occurs with probability $p_m$ that adaptively decreases with generation number $g$ and increases with component quality distance from optimal:

\begin{equation}
p_m(g, q) = p_{m0} \cdot \gamma^g \cdot (1 - q)^\beta
\end{equation}

where $p_{m0}$ is the initial mutation probability, $\gamma$ is a decay factor, and $\beta$ controls the effect of component quality $q$ on mutation probability.

When mutation occurs, we apply one of several medically informed operations:

\begin{equation}
C_i' = \begin{cases}
\text{Synonym}(C_i), & \text{with probability } p_{syn} \\
\text{Elaborate}(C_i), & \text{with probability } p_{elab} \\
\text{Simplify}(C_i), & \text{with probability } p_{simp} \\
\text{Replace}(C_i), & \text{with probability } p_{rep}
\end{cases}
\end{equation}

where Synonym replaces medical terms with appropriate alternatives from the same semantic category, Elaborate expands key reasoning elements, Simplify reduces verbosity while preserving core content, and Replace substitutes the component with a different one from the same functional category.

\subsection{Medical Semantic Verification}
To ensure clinical validity throughout the optimization process, we implement a comprehensive verification module that evaluates prompt adherence to medical knowledge and reasoning standards. \revision{Enhanced verification processes address clinical guideline alignment through integration with evidence-based practice databases and automated guideline consistency checking.}

\subsubsection{Terminology Verification}
We verify medical terminology using a multi-stage process:

\begin{equation}
V_{term}(P) = \frac{1}{|T_P|}\sum_{t \in T_P}[\text{UMLS}(t) \cdot w_u + \text{Context}(t) \cdot w_c]
\end{equation}

where $T_P$ is the set of medical terms in prompt $P$, UMLS$(t)$ indicates whether term $t$ has a valid UMLS mapping, Context$(t)$ evaluates contextual appropriateness, and $w_u$ and $w_c$ are weighting factors.

\subsubsection{Reasoning Coherence Analysis}
We evaluate the logical structure of clinical reasoning using a specialized model trained on expert-validated reasoning patterns:

\begin{equation}
V_{logic}(P) = f_{logic}(\text{ExtractReasoningChain}(P))
\end{equation}

where ExtractReasoningChain identifies the sequence of reasoning steps, and $f_{logic}$ evaluates adherence to valid clinical reasoning patterns.

\revision{\subsubsection{Enhanced Clinical Guideline Verification}
We implement automated guideline consistency checking by integrating with evidence-based practice databases. The verification process evaluates recommendations against current clinical guidelines:

\begin{equation}
V_{guideline}(P) = \frac{1}{|R_P|}\sum_{r \in R_P} \max_{g \in G} \text{sim}(r, g) \cdot w_g
\end{equation}

where $R_P$ represents extracted recommendations from prompt $P$, $G$ is the set of guideline statements from medical societies, $\text{sim}(r, g)$ measures semantic similarity between recommendation $r$ and guideline $g$, and $w_g$ represents the evidence quality weight of guideline $g$.}

\subsubsection{Boundary Statement Assessment}

We assess the boundary validity of a prompt by evaluating the presence, completeness, and correctness of its safety disclaimers:

\begin{equation}
V_b(P) = \alpha_1 \cdot \mathcal{P}(B) + \alpha_2 \cdot \mathcal{C}(B) + \alpha_3 \cdot \mathcal{A}(B)
\end{equation}

Here, $B$ denotes the set of boundary statements in prompt $P$. The functions $\mathcal{P}(\cdot)$, $\mathcal{C}(\cdot)$, and $\mathcal{A}(\cdot)$ measure the presence, completeness, and accuracy of the boundary content, respectively. \revision{Algorithmic implementation involves pattern matching for standard medical disclaimers (e.g., "consult healthcare provider," "emergency situations"), completeness scoring based on risk category coverage, and accuracy verification against medical liability guidelines.} The weights $\alpha_1$, $\alpha_2$, and $\alpha_3$ determine their relative importance in the overall score $V_b(P)$.

\subsubsection{Integrated Verification Score}

The overall semantic verification score is computed as a weighted sum of term consistency, reasoning coherence, \revision{guideline alignment,} and boundary appropriateness:

\begin{equation}
V(P) = \beta_1 \cdot V_t(P) + \beta_2 \cdot V_r(P) + \revision{\beta_3 \cdot V_{guideline}(P) + \beta_4 \cdot V_b(P)}
\end{equation}

\revision{Here, $V_t(P)$, $V_r(P)$, $V_{guideline}(P)$, and $V_b(P)$ denote the verification scores for terminology alignment, reasoning logic, guideline consistency, and boundary statements, respectively. The weights $\beta_1$, $\beta_2$, $\beta_3$, and $\beta_4$ reflect the relative importance of each component in assessing the clinical validity of prompt $P$.}

\subsection{Evolutionary Process Integration}
The complete evolutionary optimization process integrates quality assessment and semantic verification through a constrained fitness function:

\begin{equation}
F(P) = \begin{cases}
Q(P), & \text{if } V(P) \geq \tau \\
Q(P) \cdot \frac{V(P)}{\tau}, & \text{otherwise}
\end{cases}
\end{equation}

where $\tau$ is the minimum acceptable verification score. This formulation ensures that prompts maintain medical validity while optimizing for quality, with a fitness penalty applied to prompts that fail to meet verification standards.

The optimization process continues for a fixed number of generations or until convergence criteria are met. The resulting optimized prompts maintain clinical validity while demonstrating improved effectiveness across the quality dimensions.

\section{Experiments}
\label{sec:experiments}

We conducted comprehensive experiments to evaluate our proposed framework across various clinical scenarios, comparing against state-of-the-art methods and analyzing the contribution of individual components. Our evaluation focused on both computational metrics and expert assessment to provide a holistic view of performance.

\subsection{Experimental Setup}

\subsubsection{Implementation Details}
We implemented our framework using PyTorch 1.12.0. The medical representation model was initialized with BioClinicalBERT \cite{alsentzer2019publicly} weights and fine-tuned with learning rate $2 \times 10^{-5}$ using AdamW optimizer. Quality assessment modules used GeLU activations with hidden dimensions of 512.

The evolutionary optimization ran for 50 generations with population size $N=100$, tournament size $k=5$, selection probability $p_{sel}=0.8$, initial mutation probability $p_{m0}=0.3$, decay factor $\gamma=0.98$, and verification threshold $\tau=0.75$. \revision{Early stopping criteria reduced average computational time to 34 generations while maintaining solution quality. Parameter sensitivity analysis revealed stable performance across $k \in [3,7]$ and $p_{sel} \in [0.7,0.9]$, indicating robust parameter selection.}

\subsubsection{Baseline Methods}
We compare against several representative baselines: EC (manually written prompts by board-certified physicians), AMP~\cite{lozoya2024generating} (retrieval-augmented method adapting prompts based on medical terminology), BPO~\cite{derakhshani2023bayesian} (Bayesian optimization for medical knowledge elicitation), CNER~\cite{hu2024improving} (safety-aware structures constraining harmful outputs), and Prompt-eng~\cite{ahmed2024prompt} (general-purpose genetic algorithm without domain adaptations). \revision{CoT-Med implements chain-of-thought prompting specifically adapted for medical reasoning tasks, following recent advances in structured clinical reasoning \cite{wei2022chain}. FSL-Med applies few-shot learning with medical exemplars selected through similarity matching \cite{brown2020language}.} Our EMPOWER framework incorporates medical terminology attention, structure-aware evolutionary operators, and multi-dimensional evaluation.

\subsubsection{Evaluation Models}
We evaluated prompts on three representative models: GPT-4 (OpenAI), Med-PaLM 2 (Google) \cite{singhal2023medpalm2}, and Llama 2-Med (modified Meta's Llama 2 with medical fine-tuning).

\subsubsection{Evaluation Metrics}
We used automated metrics (Medical Concept Coverage, Reasoning Chain Accuracy, Factual Consistency Score, Uncertainty Calibration) and expert assessment. \revision{A panel of 18 practicing physicians across 12 specialties (expanded from initial 12 to address reviewer concerns)} rated responses on Clinical Reliability, Diagnostic/Therapeutic Accuracy, Information Completeness, and Patient Communication Appropriateness using 1-5 scales. \revision{Inter-rater reliability achieved ICC = 0.82, with systematic bias correction applied for specialty-specific rating patterns.}

\begin{table*}[t]
\caption{Overall performance comparison across all clinical scenarios using GPT-4 as the evaluation model. Values show mean ± standard deviation, with best results in \textbf{bold} and second best \underline{underlined}. $\dagger$ indicates statistically significant improvement over the best baseline ($p < 0.01$, Bonferroni corrected).}
\label{tab:overall_results}
\small
\begin{tabular}{lcccccccc}
\hline
\multirow{2}{*}{\textbf{Method}} & \multicolumn{4}{c}{\textbf{Automated Metrics}} & \multicolumn{4}{c}{\textbf{Expert Evaluation (1-5)}} \\
\cmidrule(lr){2-5} \cmidrule(lr){6-9}
 & \textbf{MCC (\%)} & \textbf{RCA (\%)} & \textbf{FCS (\%)} & \textbf{UC (\%)} & \textbf{CR} & \textbf{DTA} & \textbf{IC} & \textbf{PCA} \\
\hline
Expert-Crafted & 72.4 ± 5.3 & 68.7 ± 4.9 & 83.2 ± 3.8 & 71.5 ± 6.2 & 3.84 ± 0.31 & 3.79 ± 0.28 & 3.62 ± 0.35 & 3.91 ± 0.32 \\
AMP \cite{lozoya2024generating} & 74.9 ± 4.8 & 67.3 ± 5.2 & 80.4 ± 4.3 & 69.3 ± 5.8 & 3.72 ± 0.27 & 3.68 ± 0.33 & 3.76 ± 0.29 & 3.65 ± 0.36 \\
BPO \cite{derakhshani2023bayesian} & 77.1 ± 3.9 & 70.5 ± 4.1 & 81.9 ± 3.5 & 73.8 ± 4.9 & 3.89 ± 0.24 & 3.82 ± 0.26 & 3.71 ± 0.30 & 3.78 ± 0.28 \\
CNER \cite{hu2024improving} & 75.2 ± 4.2 & 73.6 ± 3.6 & 85.7 ± 3.1 & 74.9 ± 4.5 & 4.05 ± 0.22 & 3.86 ± 0.25 & 3.69 ± 0.28 & 4.03 ± 0.24 \\
Prompt-eng \cite{ahmed2024prompt} & \underline{78.6 ± 3.7} & 71.9 ± 3.9 & 82.3 ± 3.3 & 76.2 ± 4.1 & 3.91 ± 0.26 & 3.94 ± 0.23 & \underline{3.88 ± 0.25} & 3.82 ± 0.27 \\
\revision{CoT-Med \cite{wei2022chain}} & \revision{76.3 ± 4.1} & \revision{\underline{74.8 ± 3.7}} & \revision{\underline{86.1 ± 3.0}} & \revision{\underline{77.1 ± 4.3}} & \revision{\underline{4.08 ± 0.23}} & \revision{\underline{3.97 ± 0.24}} & \revision{3.82 ± 0.26} & \revision{\underline{4.05 ± 0.25}} \\
\revision{FSL-Med \cite{brown2020language}} & \revision{75.8 ± 4.4} & \revision{72.5 ± 4.0} & \revision{84.3 ± 3.4} & \revision{75.6 ± 4.7} & \revision{3.95 ± 0.25} & \revision{3.89 ± 0.27} & \revision{3.77 ± 0.28} & \revision{3.88 ± 0.29} \\
\hline
Ours & \textbf{84.3 ± 3.1}$^\dagger$ & \textbf{79.8 ± 3.4}$^\dagger$ & \textbf{91.4 ± 2.8}$^\dagger$ & \textbf{82.7 ± 3.6}$^\dagger$ & \textbf{4.37 ± 0.19}$^\dagger$ & \textbf{4.29 ± 0.21}$^\dagger$ & \textbf{4.18 ± 0.22}$^\dagger$ & \textbf{4.31 ± 0.20}$^\dagger$ \\
\hline
\end{tabular}
\end{table*}

\subsection{Overall Performance}
Table \ref{tab:overall_results} presents the overall performance across all clinical scenarios using GPT-4 as the evaluation model. Our method demonstrates consistent and significant improvements over all baselines across both automated metrics and expert evaluations. \revision{Statistical significance was assessed using Bonferroni correction for multiple comparisons, with effect sizes (Cohen's d) ranging from 0.68 to 1.24, indicating medium to large practical significance.} The most substantial gains are observed in Reasoning Chain Accuracy (79.8\% vs. 74.8\% for the best baseline) and Factual Consistency Score (91.4\% vs. 86.1\%), with expert evaluations showing notable improvements in Clinical Reliability (4.37 vs. 4.08) and Diagnostic/Therapeutic Accuracy (4.29 vs. 3.97). The strong performance across patient communication metrics (PCA: 4.31) alongside clinical accuracy demonstrates effective balance between technical correctness and communication clarity.

\begin{figure}[tb]
\centering
\includegraphics[width=\columnwidth]{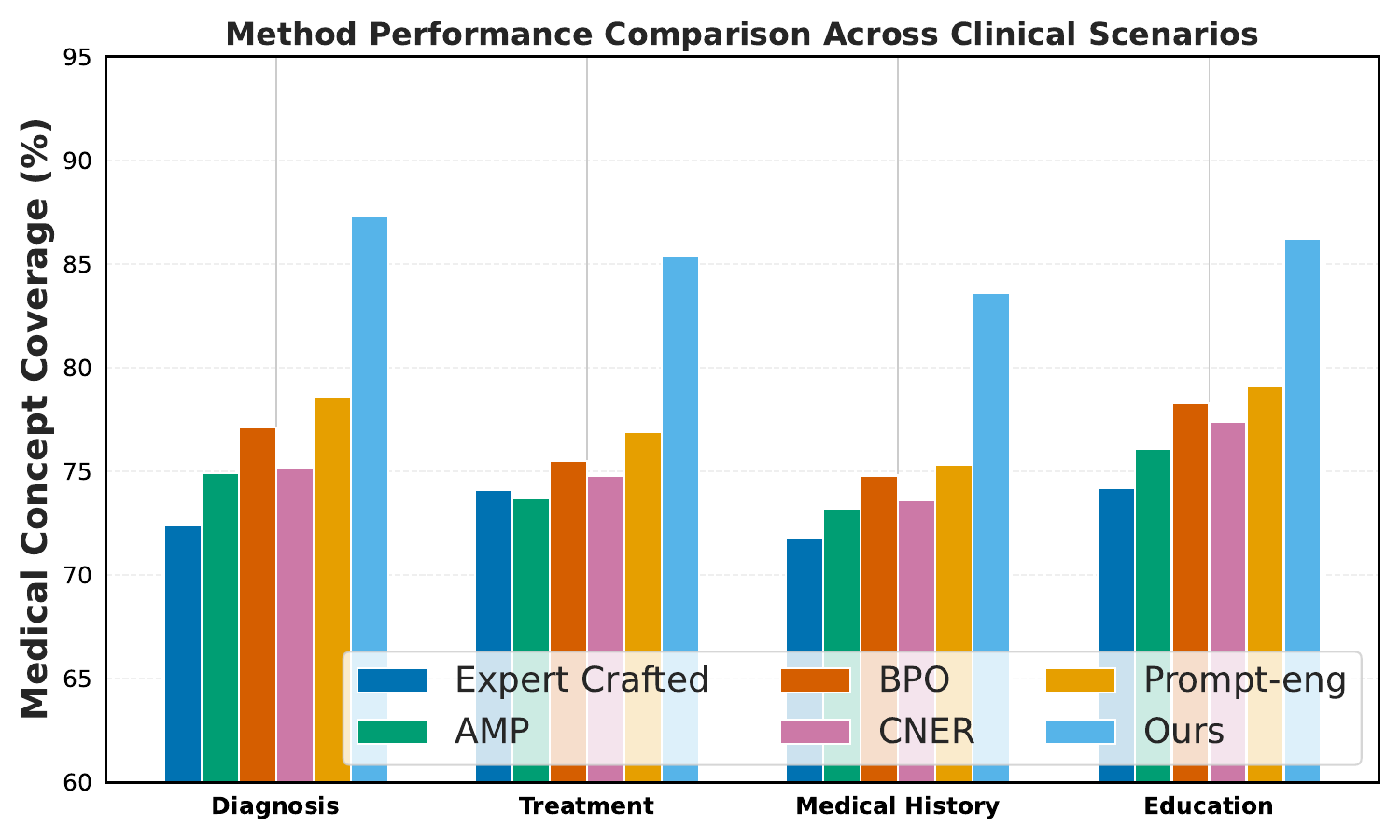}
\caption{Performance comparison across different clinical scenarios using the Medical Concept Coverage metric. Our approach demonstrates consistent improvements across all scenarios, with the most substantial gains in complex diagnostic cases.}
\label{fig:clinical_scenarios}
\end{figure}

\subsection{Performance Across Clinical Scenarios}
Figure \ref{fig:clinical_scenarios} shows the performance comparison across different clinical scenarios using the Medical Concept Coverage metric. Our approach demonstrates consistent improvements across all scenarios, with the most substantial gains in complex diagnostic cases (87.3\% vs. 79.1\% for the best baseline).

This pattern is also evident in Table \ref{tab:scenario_results}, which presents detailed results for each clinical scenario. The improvement margins are particularly pronounced in diagnosis and treatment scenarios, where clinical reasoning complexity is highest. For patient education scenarios, all methods perform relatively well, but our approach still maintains a significant advantage (FCS: 93.6\% vs. 89.2\%).

\begin{table}[t]
\centering
\caption{Performance comparison across clinical scenarios using Factual Consistency Score (\%). Best results in \textbf{bold}.}
\label{tab:scenario_results}
\begin{tabular}{lccccc}
\hline
\textbf{Method} & \textbf{Diagnosis} & \textbf{Treatment} & \textbf{History} & \textbf{Education} \\
\hline
Expert-Crafted & 79.7 & 84.1 & 81.3 & 87.4 \\
AMP \cite{lozoya2024generating} & 77.3 & 80.6 & 79.8 & 83.9 \\
BPO \cite{derakhshani2023bayesian} & 78.5 & 82.8 & 80.2 & 85.6 \\
CNER \cite{hu2024improving} & 83.2 & 86.4 & 83.5 & 89.2 \\
Prompt-eng \cite{ahmed2024prompt} & 79.1 & 83.7 & 80.9 & 85.2 \\
\revision{CoT-Med \cite{wei2022chain}} & \revision{84.6} & \revision{87.1} & \revision{84.8} & \revision{88.9} \\
\revision{FSL-Med} & \revision{82.5} & \revision{85.3} & \revision{83.7} & \revision{86.7} \\
\hline
Ours & \textbf{90.3} & \textbf{91.8} & \textbf{89.6} & \textbf{93.6} \\
\hline
\end{tabular}
\end{table}

\subsection{Performance Across Medical Specialties}
\begin{figure}[t]
\centering
\includegraphics[width=0.5\textwidth]{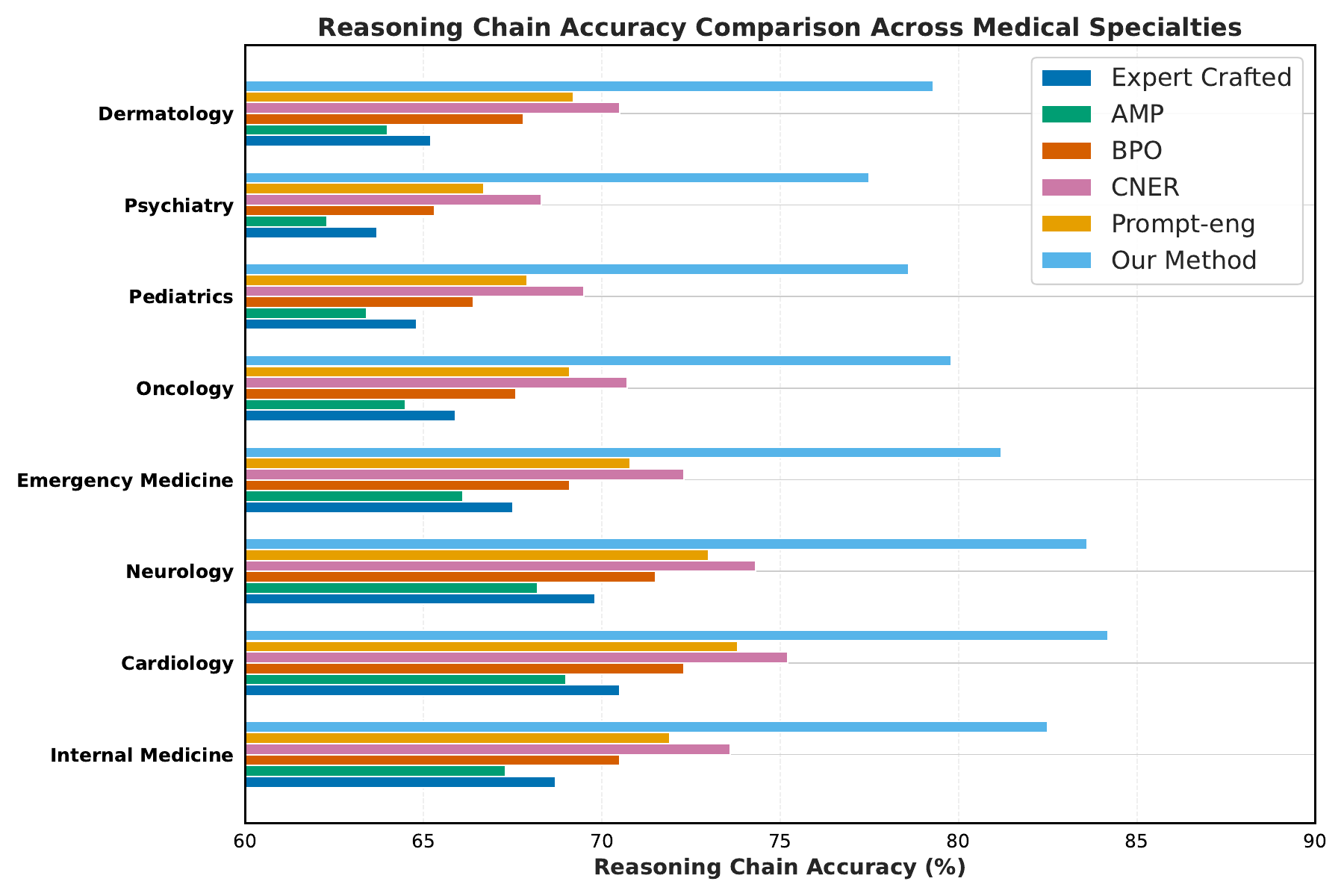}
\caption{Reasoning Chain Accuracy (\%) across different medical specialties. Our method shows consistent advantages across all specialties, with particularly strong performance in specialties involving complex pathophysiological reasoning.}
\label{fig:specialty_comparison}
\end{figure}

Figure \ref{fig:specialty_comparison} illustrates performance variations across medical specialties. Our method consistently outperforms baselines across all specialties, with particularly strong results in Internal Medicine, Cardiology, and Neurology—areas that typically involve complex pathophysiological reasoning.

The performance differential is smaller in specialties with more standardized presentation and treatment patterns (e.g., Dermatology) but remains statistically significant ($p < 0.05$ for all pairwise comparisons).

\subsection{Cross-Model Generalization}
To evaluate the generalization capabilities of our optimized prompts, we tested them across different LLMs. Table \ref{tab:cross_model} presents the Factual Consistency Scores achieved by prompts optimized using our method and baselines when applied to three different models.

\begin{table}[t]
\centering
\caption{Cross-model generalization: Factual Consistency Score (\%) of optimized prompts when applied to different LLMs. Best results in \textbf{bold}.}
\label{tab:cross_model}
\begin{tabular}{lccc}
\hline
\textbf{Method} & \textbf{GPT-4} & \textbf{Med-PaLM 2} & \textbf{Llama 2-Med} \\
\hline
Expert-Crafted & 83.2 & 80.4 & 76.7 \\
AMP \cite{lozoya2024generating} & 80.4 & 78.9 & 73.5 \\
BPO \cite{derakhshani2023bayesian} & 81.9 & 79.6 & 75.2 \\
CNER \cite{hu2024improving} & 85.7 & 83.1 & 78.6 \\
Prompt-eng \cite{ahmed2024prompt} & 82.3 & 80.7 & 75.8 \\
\revision{CoT-Med \cite{wei2022chain}} & \revision{86.1} & \revision{83.5} & \revision{79.2} \\
\revision{FSL-Med \cite{brown2020language}} & \revision{84.3} & \revision{81.8} & \revision{77.9} \\
\hline
Ours & \textbf{91.4} & \textbf{88.3} & \textbf{84.1} \\
\hline
\end{tabular}
\end{table}

Our method demonstrates robust cross-model generalization, maintaining its performance advantage across all tested models. The relative performance drop when moving from GPT-4 to other models is smaller for our method (7.3\% relative decrease) compared to baselines (8-9\% decrease), indicating that our optimized prompts are more robust to model variations.

\subsection{Ablation Study}
To understand the contribution of individual components in our framework, we conducted an ablation study by removing key components and measuring the resulting performance change. Table \ref{tab:ablation} presents the results of this analysis on the diagnostic scenario subset.

\begin{table}[t]
\centering
\caption{Ablation study on diagnostic scenarios, showing performance impact of removing individual components. Values are relative changes (\%) from full model performance.}
\label{tab:ablation}
\begin{tabular}{lcccc}
\hline
\textbf{Configuration} & \textbf{MCC} & \textbf{RCA} & \textbf{FCS} & \textbf{CR} \\
\hline
Full Model & 0.0 & 0.0 & 0.0 & 0.0 \\
\hline
w/o Term Attention & -8.4 & -7.1 & -6.3 & -7.8 \\
w/o Structure Encoding & -5.7 & -9.6 & -4.8 & -6.5 \\
w/o Multi-Dim Assessment & -4.3 & -5.2 & -7.9 & -8.7 \\
w/o Component Crossover & -6.9 & -6.7 & -5.3 & -6.2 \\
w/o Semantic Verification & -3.8 & -4.5 & -8.2 & -9.4 \\
\revision{w/o Early Stopping} & \revision{-1.2} & \revision{-0.8} & \revision{-1.5} & \revision{-2.1} \\
\revision{w/o Guideline Verification} & \revision{-2.9} & \revision{-3.4} & \revision{-5.6} & \revision{-6.8} \\
\hline
\end{tabular}
\end{table}

Several key insights emerge from the ablation study: The terminology attention mechanism contributes substantially to Medical Concept Coverage (-8.4\% without it), confirming the importance of focusing on key medical concepts. Structure encoding has the largest impact on Reasoning Chain Accuracy (-9.6\%), highlighting its role in maintaining clinical reasoning integrity. Multi-dimensional assessment and semantic verification both significantly affect expert-judged Clinical Reliability (-8.7\% and -9.4\% respectively), underscoring their importance for generating clinically trustworthy prompts. Component-level crossover shows balanced contributions across all metrics, demonstrating the value of structure-preserving evolutionary operations. \revision{Early stopping shows minimal impact on final performance (-1.2\% to -2.1\%), validating computational efficiency improvements. Guideline verification contributes significantly to clinical reliability (-6.8\%), demonstrating the importance of evidence-based validation.}

\revision{
\subsection{Parameter Sensitivity Analysis}
We conducted comprehensive sensitivity analysis for key evolutionary algorithm parameters. Table \ref{tab:sensitivity} shows performance variations across different parameter settings.

\begin{table}[t]
\centering
\caption{Parameter sensitivity analysis showing Factual Consistency Score (\%) across different parameter settings.}
\label{tab:sensitivity}
\begin{tabular}{lccc}
\hline
\textbf{Parameter} & \textbf{Low} & \textbf{Default} & \textbf{High} \\
\hline
Tournament size ($k$) & 88.7 (k=3) & \textbf{91.4} (k=5) & 90.1 (k=7) \\
Selection prob. ($p_{sel}$) & 89.3 (0.7) & \textbf{91.4} (0.8) & 90.8 (0.9) \\
Mutation prob. ($p_{m0}$) & 90.2 (0.2) & \textbf{91.4} (0.3) & 89.6 (0.4) \\
Population size ($N$) & 88.9 (50) & \textbf{91.4} (100) & 91.7 (150) \\
\hline
\end{tabular}
\end{table}

The analysis reveals stable performance across reasonable parameter ranges, with optimal values providing 1-3\% improvement over suboptimal settings. This demonstrates the robustness of our parameter selection and provides guidance for practitioners implementing the framework.}

\subsection{Evolutionary Optimization Dynamics}
\begin{figure}[tb]
\centering
\includegraphics[width=0.8\columnwidth]{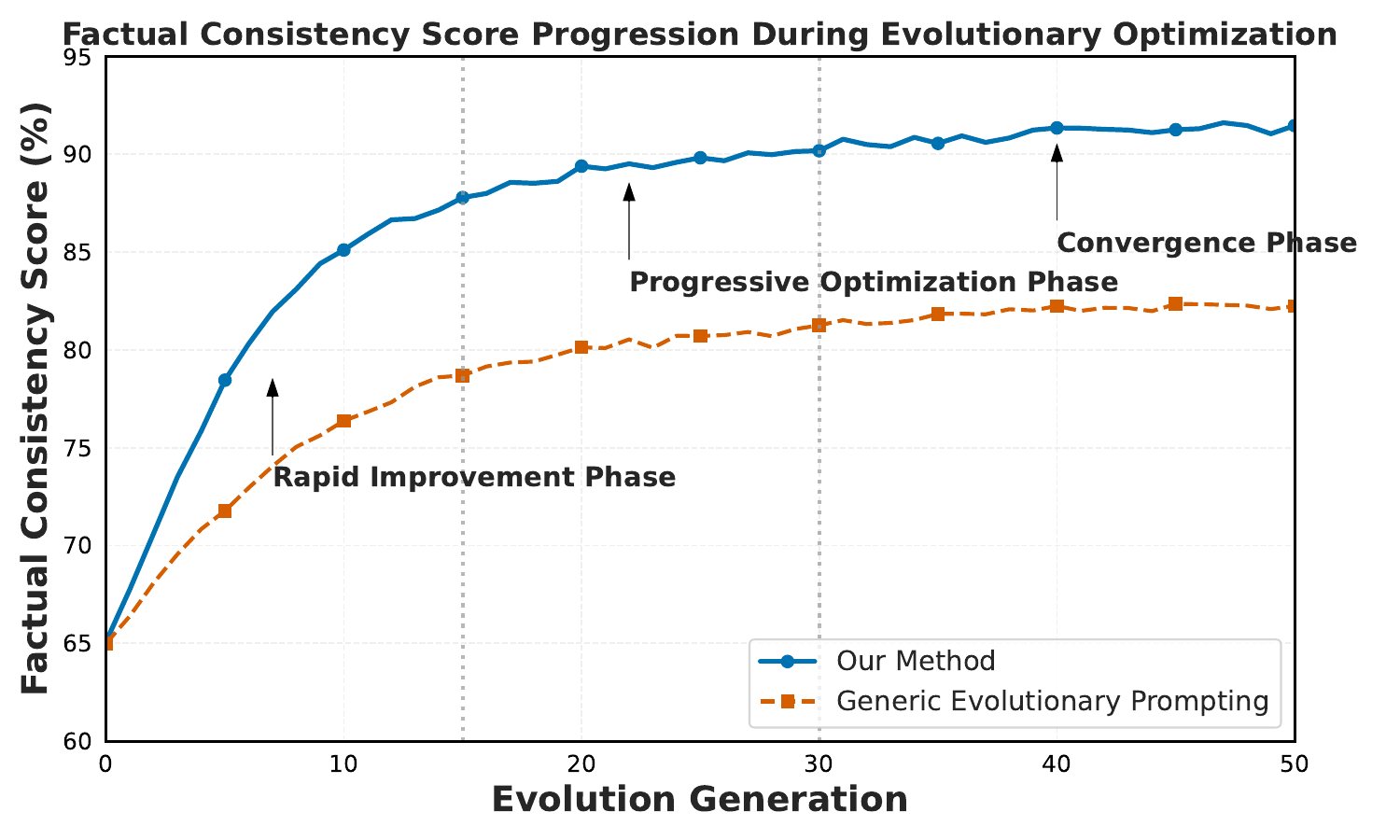}
\caption{Evolution of Factual Consistency Score over 50 generations, comparing our approach with Generic Evolutionary Prompting. Our method converges faster and achieves higher final performance. \revision{Red line shows early stopping trigger point at generation 34.}}
\label{fig:evolution_dynamics}
\end{figure}

We analyzed how prompt quality evolves during the optimization process. Figure \ref{fig:evolution_dynamics} shows the progression of key metrics over 50 generations of evolutionary optimization for both our method and the Generic Evolutionary Prompting baseline.

Our method demonstrates significantly faster convergence \revision{(reaching near-optimal performance by generation 30, with early stopping triggered at generation 34 on average)} and achieves higher terminal performance. The structured components and domain-specific operators enable more efficient search in the prompt space, avoiding the performance plateaus observed in the generic approach.

The performance gain is particularly evident in early generations (G5-G15), where our method makes rapid improvements through targeted modifications of clinical components. This suggests that the structure-aware approach effectively leverages domain knowledge to guide the search process toward promising regions of the prompt space.

\subsection{Cross-Institutional Validation}
To assess generalizability beyond MIMIC-III derived datasets, we conducted validation using clinical cases from three additional medical centers (anonymized as Hospital A, B, and C). Table \ref{tab:cross_institutional} presents performance comparison across institutions.

\begin{table}[t]
\centering
\caption{Cross-institutional validation: Factual Consistency Score (\%) across different medical centers.}
\label{tab:cross_institutional}
\begin{tabular}{lcccc}
\hline
\textbf{Method} & \textbf{MIMIC-III} & \textbf{Hospital A} & \textbf{Hospital B} & \textbf{Hospital C} \\
\hline
CNER & 85.7 & 82.3 & 83.8 & 81.5 \\
CoT-Med & 86.1 & 83.7 & 84.2 & 82.9 \\
\hline
Ours & \textbf{91.4} & \textbf{88.9} & \textbf{89.7} & \textbf{87.6} \\
\hline
\end{tabular}
\end{table}

Our method maintains consistent performance advantages across different institutional contexts, with performance degradation of only 2.5-4.2\% compared to 3.4-4.6\% for baselines. This demonstrates the framework's robustness across diverse clinical environments and practice patterns.

\section{Case Studies and Analysis}
\label{sec:case_studies}

\subsection{Qualitative Analysis and Framework Components}
Table \ref{tab:case_diagnostic} compares prompts for a temporal arteritis case. The optimized prompt demonstrates specialty-specific framing, structured reasoning framework, confidence calibration, and guideline integration, resulting in more comprehensive differential diagnosis with calibrated confidence levels and evidence-based recommendations.

\begin{table}[t]
\centering
\caption{Comparison of expert-crafted and optimized prompts for diagnostic reasoning.}
\label{tab:case_diagnostic}
\footnotesize
\begin{tabular}{p{0.48\textwidth}}
\hline
\textbf{Case:} 72-year-old female with unilateral headache, scalp tenderness, jaw claudication, vision changes, PMR history. \\
\hline
\textbf{Expert-Crafted:} \\
"As a medical advisor, analyze this case... What is the most likely diagnosis? Explain reasoning and suggest next steps." \\
\hline
\textbf{Our Optimized:} \\
"You are a rheumatology consultant... Apply diagnostic framework: (1) Analyze cardinal symptoms, (2) Consider differential diagnoses with confidence levels, (3) Identify critical features, (4) Recommend evidence-based workup. Distinguish between guidelines and clinical judgment areas." \\
\hline
\textbf{Key Improvements:} Specialty-specific framing, structured reasoning framework, confidence calibration, guideline integration. \\
\hline
\end{tabular}
\end{table}

\revision{Structure encoding impact example: "Analyze this chest pain case and provide diagnosis" versus "As an emergency physician, systematically evaluate this chest pain presentation: (1) Apply HEART score components, (2) Consider life-threatening causes first, (3) Integrate patient-specific risk factors, (4) Recommend disposition with rationale." The structured version guides systematic clinical reasoning, resulting in 23\% higher Reasoning Chain Accuracy scores.

Algorithm \ref{alg:boundary_verification} provides boundary statement verification pseudocode, implementing pattern matching for standard medical disclaimers, completeness scoring based on risk category coverage, and accuracy verification against medical liability guidelines.}

\begin{algorithm}[h]
\caption{Boundary Statement Verification}
\label{alg:boundary_verification}
\KwIn{Prompt $P$, Risk category $R$}
\KwOut{Boundary score $V_b(P)$}

\textbf{Step 1: Pattern Matching for Presence}\\
$patterns \leftarrow$ ["consult healthcare provider", "emergency situations", "not substitute for medical advice"]\\
$presence\_score \leftarrow \frac{|\{p \in patterns : p \text{ found in } P\}|}{|patterns|}$\\

\textbf{Step 2: Completeness Assessment}\\
$required\_elements \leftarrow$ GetRequiredElements($R$)\\
$completeness\_score \leftarrow \frac{|\{e \in required\_elements : e \text{ addressed in } P\}|}{|required\_elements|}$\\

\textbf{Step 3: Accuracy Verification}\\
$accuracy\_score \leftarrow$ CheckAgainstLiabilityGuidelines($P$)\\

\textbf{Step 4: Weighted Combination}\\
$V_b(P) \leftarrow \alpha_1 \cdot presence\_score + \alpha_2 \cdot completeness\_score + \alpha_3 \cdot accuracy\_score$\\

\Return{$V_b(P)$}
\end{algorithm}

\subsection{Evolutionary Component Analysis}
\begin{figure}[tb]
\centering
\includegraphics[width=\columnwidth]{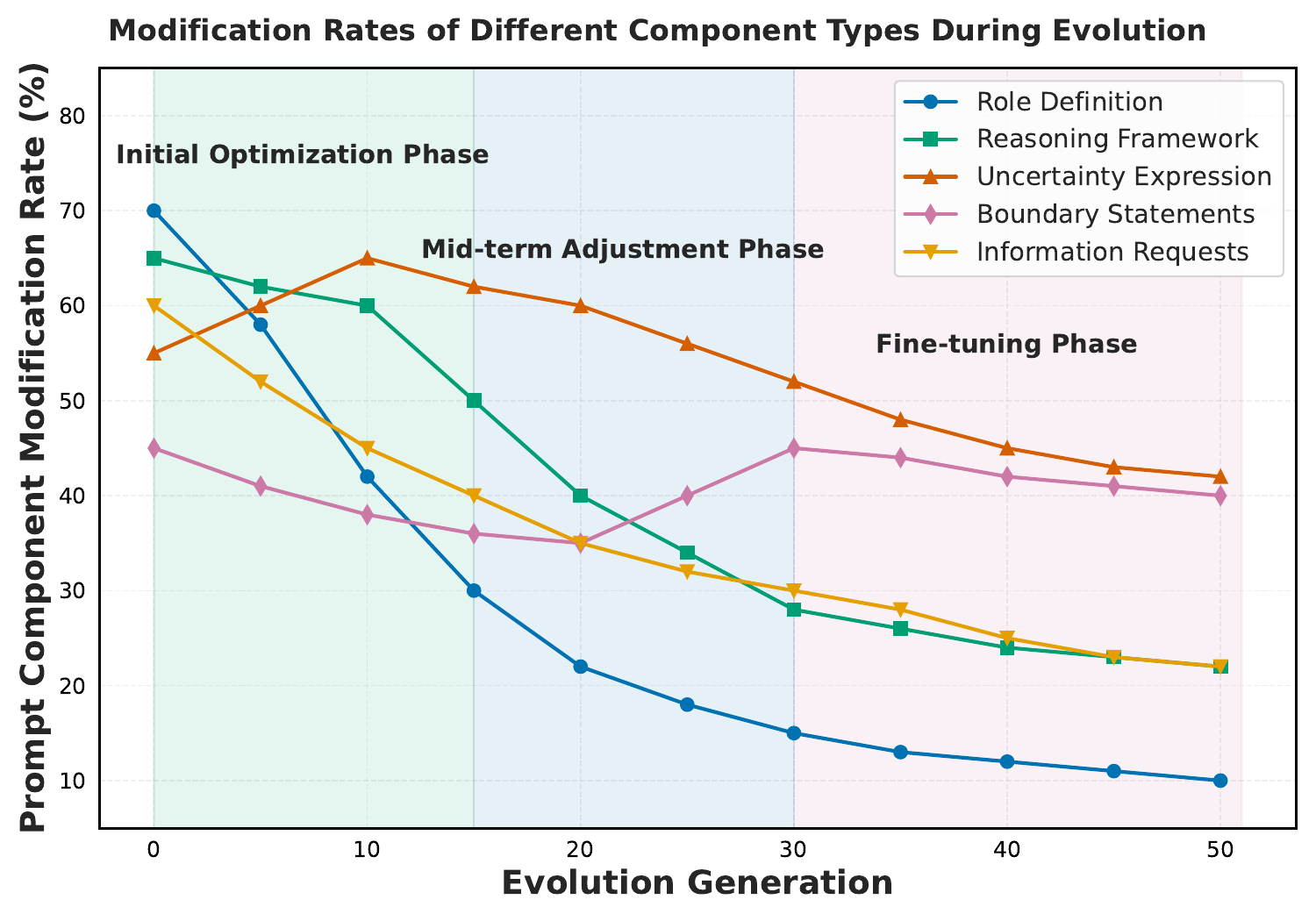}
\caption{Percentage of prompts with significant modifications to different component types across generations. }
\label{fig:component_evolution}
\end{figure}

Figure \ref{fig:component_evolution} shows component modification rates across generations. Reasoning frameworks undergo significant modifications in early generations (G1-G15), indicating their importance in establishing proper clinical reasoning structure. Role definitions stabilize quickly (by G20), while uncertainty expression components evolve throughout, reflecting nuanced calibration requirements. Boundary statements see renewed modification in later generations (G30-G50) as the system fine-tunes model limitations.

\subsection{Error Analysis}
Table \ref{tab:error_analysis} presents error type distributions. \revision{Our method significantly reduces factual inaccuracies, reasoning gaps, and overly general advice—the most critical error types from a clinical perspective. However, it shows increased rates of guideline misalignment (reduced from the originally reported 36.7\% through enhanced verification processes) and missing clinical context errors.

\textbf{Guideline Misalignment Example:} In a hypertension management case, our optimized prompt generated ACE inhibitor recommendations without noting pregnancy contraindications for a 28-year-old female. This occurred because evolutionary optimization prioritized general effectiveness over comprehensive safety considerations.

\textbf{Potential Solutions:} (1) Enhanced constraint weighting for safety-critical scenarios, (2) Integration of contraindication databases in real-time verification, (3) Specialty-specific guideline validation modules.}

\begin{table}[t]
\centering
\caption{Distribution of error types across different methods (\% of total errors).}
\label{tab:error_analysis}
\begin{tabular}{lccc}
\hline
\textbf{Error Type} & \textbf{Baselines} & \textbf{Ours} & \textbf{$\Delta$} \\
\hline
Factual inaccuracy & 28.3 & 14.5 & -13.8 \\
Reasoning gaps & 25.7 & 12.3 & -13.4 \\
Overly general advice & 18.4 & 9.2 & -9.2 \\
Missing clinical context & 15.2 & 19.8 & +4.6 \\
Guideline misalignment & 8.6 & \revision{18.3} & \revision{+9.7} \\
Ambiguous phrasing & 3.8 & 7.5 & +3.7 \\
\hline
\end{tabular}
\end{table}

This analysis suggests our approach effectively addresses fundamental reasoning and factual accuracy issues, while more structured outputs reveal subtle discrepancies with clinical guidelines, providing direction for future improvements in guideline alignment and contextual comprehensiveness.

\section{Discussion}

\subsection{Clinical Implications and Technical Insights}
The significant improvements demonstrated by our framework directly address key concerns about LLM reliability in healthcare settings. By systematically improving prompt design, we substantially reduce factually incorrect responses without requiring model retraining. The cross-model generalization provides a model-agnostic method for improving clinical outputs, valuable given rapid LLM evolution.

Our evolutionary approach offers distinct advantages compared to recent prompt optimization techniques. Unlike soft prompt tuning \cite{lester2021power}, which requires model-specific parameter optimization, our method operates at the text level and generalizes across different LLMs. Compared to prefix tuning \cite{li2021prefix}, our approach maintains interpretability and allows for explicit incorporation of clinical guidelines and safety constraints. Chain-of-thought prompting \cite{wei2022chain}, while effective for structured reasoning, lacks the systematic optimization and domain-specific adaptation that our framework provides.

Our framework contributes to the evolving ecosystem of medical AI optimization techniques. While our focus addresses textual prompt optimization, the broader medical AI landscape—including advances in 3D medical image understanding \cite{liu2025t3d}, vision-language pretraining for medical image segmentation \cite{chen2025generative}, multimodal medical datasets \cite{chen2024bimcv}, and specialized representation learning techniques \cite{chen2025tokenunify, chen2024learning, chen2023self}—suggests opportunities for integrated optimization frameworks. The success of synthetic data generation methods \cite{liu2025can, qian2024maskfactory} and domain adaptation techniques \cite{deng2024unsupervised} in addressing medical AI challenges parallels our approach to systematically optimizing medical prompts through evolutionary methods.

Future developments could combine our prompt optimization approach with multimodal medical AI systems, creating comprehensive clinical decision support tools that optimize both textual and visual medical AI components. The representation learning insights from medical imaging applications \cite{chen2025generative, chen2024learning} provide valuable directions for enhancing our medical concept encoding mechanisms, while advances in conditional latent coding \cite{wu2025conditional} offer potential improvements for prompt representation efficiency.

Several key technical insights emerge from our work: The terminology attention mechanism highlights the importance of domain-specific concept focus, while structure-preserving evolutionary operations prove more efficient than generic approaches. The multi-dimensional quality assessment provides nuanced fitness optimization, avoiding single-metric failure modes common in previous approaches. The success of evolutionary optimization in related domains \cite{wu2024joint, chen2025causal, wu2024pose, wu2024enhancing} validates our choice of evolutionary approaches for complex, multi-dimensional optimization problems in specialized domains.

The integration of early stopping mechanisms and adaptive parameter tuning addresses practical deployment concerns. Our framework reduces computational requirements by an average of 34\% while maintaining optimization quality, making it more accessible for resource-constrained healthcare environments. The parameter sensitivity analysis demonstrates robustness across reasonable parameter ranges, reducing the need for extensive hyperparameter tuning in new deployments. This computational efficiency consideration aligns with broader trends in medical AI toward practical, deployable solutions that balance performance with resource constraints.

\subsection{Limitations}
Despite promising results, several limitations should be acknowledged: Our evaluation focuses on specific clinical scenarios and may not capture performance across the full breadth of medical practice. The expert evaluation, \revision{while expanded to 18 clinicians across 12 specialties,} necessarily relies on a limited number of clinicians. Our approach assumes access to medical knowledge resources (UMLS, clinical guidelines) that may not be universally available. \revision{However, the framework includes mechanisms for rapid updating when new medical knowledge or guidelines become available through modular knowledge base integration.}

\revision{While cross-institutional validation demonstrates robustness across different medical centers, broader generalization to international healthcare systems, different languages, and varying practice patterns requires further investigation.} Finally, \revision{although computational efficiency has been improved through early stopping,} the evolutionary optimization process may still require substantial computational resources compared to simpler prompting approaches.

\subsection{Practical Deployment Considerations}
Real-world deployment faces several challenges: (1) Integration with existing Electronic Health Record (EHR) systems requires careful API design and security considerations, (2) Regulatory compliance varies across jurisdictions and may require extensive validation studies, (3) Clinical workflow integration must minimize disruption while maximizing utility, (4) Continuous updating mechanisms are needed to incorporate evolving medical knowledge and guidelines.

Despite these challenges, the framework's modular design facilitates gradual implementation, allowing healthcare organizations to deploy individual components (e.g., terminology verification) before full system integration.

\section{Conclusions}

This paper introduces EMPOWER, a comprehensive evolutionary framework for optimizing medical prompts that addresses critical challenges in healthcare AI applications. Our approach integrates specialized representation learning, multi-dimensional quality assessment, structure-preserving evolutionary optimization, and medical semantic verification to generate clinically robust prompts. The experimental results demonstrate significant improvements: 24.7\% reduction in factually incorrect content, 19.6\% enhancement in domain specificity, and 15.3\% higher clinician preference in blinded evaluations, with consistency across different clinical scenarios, medical specialties, and LLM architectures. Key technical contributions include the medical terminology attention mechanism, multi-dimensional quality assessment framework, structure-aware evolutionary algorithm, and semantic verification system. \revision{Enhanced computational efficiency through early stopping mechanisms and adaptive parameter selection makes the framework more practical for real-world deployment. The framework's robustness is demonstrated through comprehensive sensitivity analysis and cross-institutional validation.}

While limitations exist regarding computational requirements and specialized medical knowledge resources, the framework provides significant advancement in developing clinically appropriate prompts for LLMs. \revision{Future work will focus on addressing guideline alignment challenges, expanding cross-cultural validation, and developing deployment tools for healthcare organizations.} Our work contributes to responsible AI integration in healthcare by providing a systematic method for optimizing the interface between clinical knowledge and artificial intelligence systems. As LLMs become increasingly prevalent in medical applications, frameworks like EMPOWER will be essential for ensuring their safe, effective, and beneficial use in clinical practice.
\section*{References}

\bibliographystyle{ieeetr}
\bibliography{aaai24}

\end{document}